%% file: main.tex
\pdfoutput=1
\documentclass[11pt,a4paper]{article}
\usepackage[hyperref]{emnlp2020}
\usepackage{times}
\usepackage{latexsym}

\usepackage{microtype}

\usepackage{graphicx}  
\usepackage{xspace}
\usepackage{eufrak}
\usepackage{amsfonts}
\usepackage{amssymb}
\usepackage{amsmath}
\usepackage{bm}
\usepackage{multirow}
\usepackage{booktabs}
\usepackage{subfig}
\usepackage[utf8]{inputenc}

\usepackage{cleveref}

\crefname{section}{§}{§§}
\Crefname{section}{§}{§§}

\newcommand{\ie}{\emph{i.e.}\xspace} 
\newcommand{\eg}{\emph{e.g.}\xspace} 

\newcommand{\x}{\bm{x}}
\newcommand{\y}{\bm{y}}
\newcommand{\s}{\bm{s}}

\newcommand{\R}{\mathbb{R}}

\newcommand{\tsrc}{\bm{s}_{\rm src}}
\newcommand{\ttgt}{\bm{s}_{\rm tgt}}
\newcommand{\spred}{\hat{\bm{x}}_{\rm tgt}}
\newcommand{\stgt}{\bm{x}_{\rm tgt}}
\newcommand{\stmpl}{\bm{x}_{\rm tmpl}}
\newcommand{\ssrc}{\bm{x}_{\rm src}}

\newcommand{\uscr}[2]{#1^{(#2)}}

\newcommand{\ours}{{GuiG}\xspace}
\newcommand{\synexpan}{{\ours.SE}\xspace}
\newcommand{\guigen}{{\ours.TG}\xspace}

\aclfinalcopy 

\title{Transformer-Based Neural Text Generation with Syntactic Guidance}

\author{Yinghao Li \\
  Georgia Institute of Technology \\
  \texttt{yinghaoli@gatech.edu} \\\And
  Rui Feng \\
  Georgia Institute of Technology \\
  \texttt{rfeng@gatech.edu} \\\AND
  Isaac Rehg \\
  Georgia Institute of Technology \\
  \texttt{isaacrehg@gatech.edu} \\\And
  Chao Zhang \\
  Georgia Institute of Technology \\
  \texttt{chaozhang@gatech.edu} \\}

\date{}

\begin{document}
\maketitle

\input{0-abstract.tex}

\input{1-introduction.tex}

\input{2-problem-setup.tex}

\input{3-methodology.tex}

\input{4-experiments.tex}

\input{5-related-works.tex}

\input{6-conclusion.tex}

\bibliography{references}
\bibliographystyle{acl_natbib}



\end{document}

%% file: 0-abstract.tex
\begin{abstract}

We study the problem of using (partial) constituency parse trees as syntactic guidance for controlled text generation.
Existing approaches to this problem use recurrent structures, which not only
suffer from the long-term dependency problem but also falls short in modeling the tree structure of the syntactic guidance.
We propose to leverage the parallelism of Transformer to better incorporate parse trees.
Our method first expands a partial template constituency parse tree to a full-fledged parse tree tailored for the input source text, and then uses the expanded tree to guide text generation.
The effectiveness of our model in this process hinges upon two new attention mechanisms:
1) a \textit{path attention} mechanism that forces one node to attend to only other nodes located in its path in the syntax tree to better incorporate syntax guidance;
2) a \textit{multi-encoder attention} mechanism that allows the decoder to dynamically attend to information from multiple encoders.
Our experiments in the controlled paraphrasing task show that our method
outperforms SOTA models both semantically and syntactically, improving the best
baseline's BLEU score from $11.83$ to $26.27$.

\end{abstract}

%% file: 1-introduction.tex
\section{Introduction}

Generating text that conforms to syntactic or semantic constraints benefits many NLP applications.
To name a few, when paired data are limited, \citet{yang-etal-2019-low} build templates from large-scale unpaired data to aid the training of the dialog generation model;
\citet{Niu2017ASO} and \citet{liu-etal-2019-rhetorically} apply style constraints to adjust the formality or rhetoric of the utterances;
\citet{iyyer2018adversarial} and \citet{li-etal-2019-Insufficient} augment dataset using controlled generation to improve the model performance.

\begin{figure}[tbp]
  \centerline{\includegraphics[width = 0.48\textwidth]{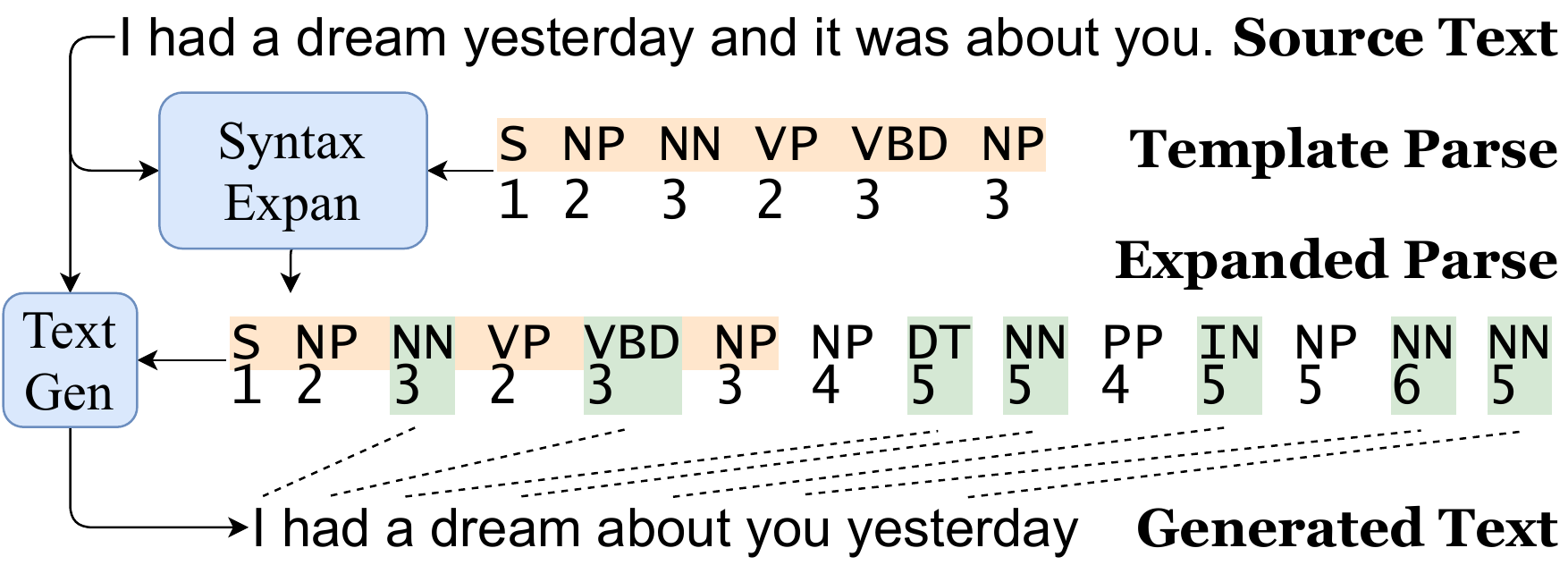}}
  \caption{The pipeline of the syntactically guided paraphrasing process.
  The template parse is the top-$\ell$ levels of a full-fledged parse tree.
  It is first expanded to a full tree, with which the target text is generated to match the semantics of the source text.
  }
  \label{fig:pipeline}
\end{figure}

We study the problem of syntactically controlled text generation, which aims to generate target text with pre-defined syntactic guidance.
Most recent studies on this topic \citep{chen2019controllable, bao2019generating} use sentences as exemplars to specify syntactic guidance.
However, the guidance specified by a sentence can be vague, because its syntactic and semantic factors are tangled.
Different from them, we use constituency parse trees as \emph{explicit} syntactic constraints.
As providing full-fledged parse trees of the target text is impractical, we require only a template parse tree that sketches a few top levels of a full tree (\cref{sec:problem-setup}).
Figure~\ref{fig:pipeline} shows our pipeline.

\citet{iyyer2018adversarial} adopt the same setting as ours.
Their proposed SCPN model uses two LSTM \citep{hochreiter1997long} encoders to respectively encode source text and parse tree, and connects them to one decoder with additional attention \citep{bahdanau2014neural} and pointer \citep{see2017get} structures.
Nonetheless, recurrent encoders not only suffer from information loss by compressing a whole sequence into one vector but also are incapable of properly modeling the tree structure of constituency parse as well.
Consequently, their network tends to ``translate'' the parse tree, instead of learning the real syntactic structures from it.

We propose a Transformer-based syntax-guided text generation method, named \ours.
It first expands a template constituency parse tree to a full-fledged parse tree tailored for the input source text, and then uses the full tree to guide text generation.
To capture the tree structure of the syntax, we apply a \emph{path attention} mechanism (\cref{subsec:text-generation}) to our text generation model.
It forces one node to attend to only other nodes located in its path (\ie, its ancestors and descendants) instead of all the nodes in the tree.
Such a mechanism limits the information flow among the nodes in the constituency tree that do not have the direct ancestor-descendant relationship, forcing the parent nodes to carry more information than their children.
In cooperation with path attention, we linearize the constituency trees to a more compact \emph{node-level} format (\cref{subsec:syntax-expansion}).
Moreover, to address the challenge of properly integrating the semantic and syntactic information, we design a \emph{multi-encoder attention} mechanism (\cref{subsec:syntax-expansion}).
It enables the Transformer decoder to accept outputs from multiple encoders simultaneously.

We evaluated our model on the controlled paraphrasing task.
The experiment results show that \ours outperforms the state-of-the-art SCPN method by $6.7\%$ in syntactic quality and $122.1\%$ in semantic quality.
Human evaluations prove our method generates  semantically and syntactically superior sentences, with $1.13$ semantic and $0.62$ syntactic score improvements.
Further, we find that the multi-encoder attention mechanism enhances the Transformer's ability to deal with multiple inputs, and the path attention mechanism significantly contributes to the model's semantic performance (\cref{sec:experiments}).

Our contributions include:
1) a multi-encoder attention mechanism that allows a Transformer decoder to attend to multiple encoders;
2) a path attention mechanism designed to better incorporate tree-structured syntax guidance with a special tree linearization format; and
3) a syntax-guided text generation method \ours that achieves new state-of-the-art semantic and syntactic performance.

%% file: 2-problem-setup.tex
\section{Problem Setup}
\label{sec:problem-setup}


Syntax-guided text generation aims to generate target text $\ttgt$ from 1) a source sentence $\tsrc$ and 2) a syntax template $\stmpl$, such that the generated sentence utilizes the semantics of $\tsrc$ and is syntactically aligned with $\stmpl$.

For the sentences, we tokenize them into sub-word units using \textit{byte pair encoding} (BPE) \cite{Sennrich_2016}.
This prevailing encoding method not only solves the out-of-vocabulary (OOV) issue but also has the ability to model the character of word roots and affixes.
Formally, the tokenized text sequence is represented by $\s = (\uscr{s}{1}, \uscr{s}{2}, \dots, \uscr{s}{M})$ with $\uscr{s}{i} \in \mathcal{C}$, where $\mathcal{C}$ is the set of all sub-word units and $M$ is the sequence length.
Moreover, we assume the constituency parse tree of the source sentence $\tsrc$ is also available, denoted as the \emph{source parse} $\ssrc$.

The syntax template $\stmpl$ is a partial constituency parse tree that provides high-level syntax sketches.
We use the top-$\ell$ ($\ell=3$ in this work) levels of \emph{target parse} $\stgt$, which is the full-fledged constituency tree of $\ttgt$.
$\stmpl$ can be also frequent templates mined from any text corpora.




%% file: 3-methodology.tex
\section{Methodology}

Our method \ours contains two models (Figure~\ref{fig:pipeline})---a syntax expander (\cref{subsec:syntax-expansion}) that expands the template parse, and a text generator (\cref{subsec:text-generation}) that leverages the expanded parse to control text generation.

\begin{figure*}[tbp]
    \centerline{\includegraphics[width = 0.95\textwidth]{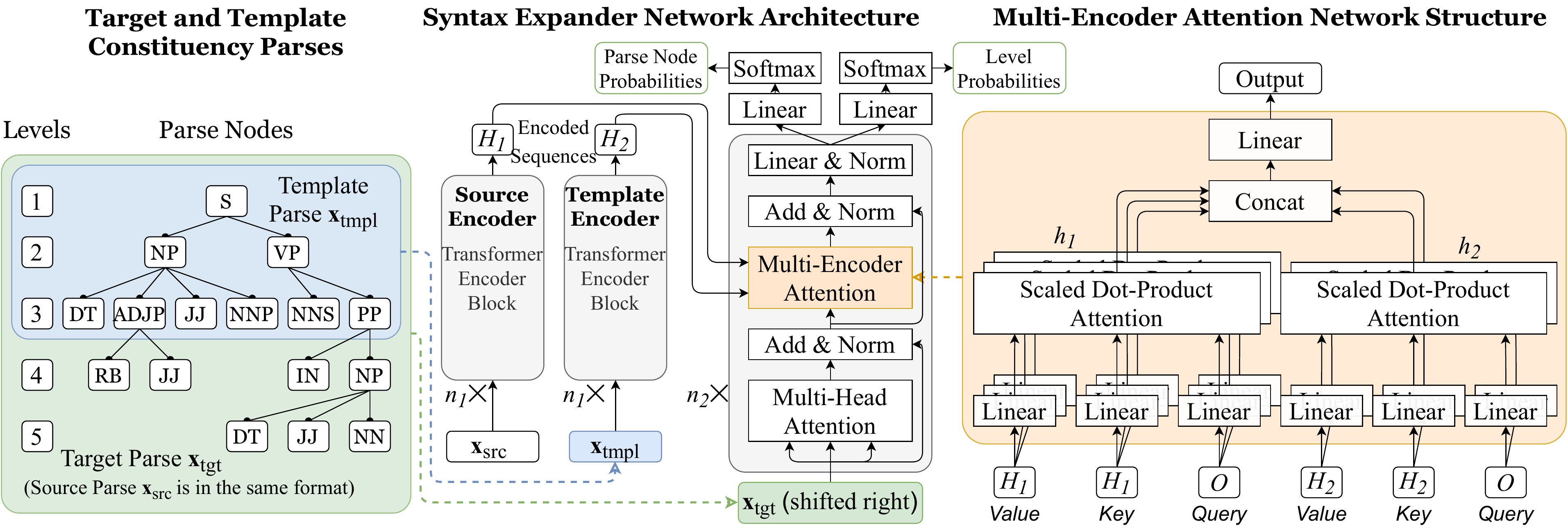}}
    \caption{
    The left block is the syntactic representation of the sentence ``{\fontfamily{qcr}\selectfont the very quick brown fox jumps over the lazy dog}'' with template parse truncation depth $\ell = 3$;
    in the middle is the model architecture of the syntax expander, and on the right is the detailed structure of the multi-encoder attention mechanism.
    $n_1$ and $n_2$ are the number of blocks;
    $h_1$ and $h_2$ are the number of attention heads;
    ``$O$'' indicates the output of the previous network layer.
    }
    \label{fig:syntax-expansion}
\end{figure*}

\subsection{Syntax Expansion}
\label{subsec:syntax-expansion}



The goal of our syntax expander is to construct a valid full-fledged target parse tree $\spred$ from the template parse $\stmpl$.
To adapt $\spred$ to the source text $\tsrc$, we use the source parse $\ssrc$ of $\tsrc$ to guide the syntax expansion process.

\paragraph{Parse Tree Linearization}

We use a pair of \textit{node} and \textit{level} sequences to represent the constituency parse tree.
A constituency parse tree $\x$ is thus linearized to a \textit{node-level} format sequence $\x = (\uscr{x}{1}, \uscr{x}{2}, \dots, \uscr{x}{N})$ where $N$ is the number of nodes in the parse tree $\x$.
For each $\uscr{x}{i} = \{ \uscr{p}{i}, \uscr{l}{i} \}$, $\uscr{p}{i}$ is the parse node and $\uscr{l}{i}$ is its level.
For example, the parse of the sentence ``{\fontfamily{qcr}I ate an apple}'' is represented by the node sequence ``{\fontfamily{qcr}\selectfont S NP PRP VP VBD NP DT NN}'' and the level sequence ``{\fontfamily{qcr}\selectfont1 2 3 2 3 3 4 4}''.
Comparing with the existing \textit{bracketed} format, which linearizes the above sentence to ``{\fontfamily{qcr}\selectfont(S(NP(PRP))(VP(VBD)(NP(DT)(NN))))}'', our node-level representation reduces the parse sequence length to $1/3$.
This more compact representation decreases the time consumption for both syntax encoding and prediction, thus facilitating the syntax expansion and text generation steps.

At the embedding layer, the parse node tokens and level tokens are embedded respectively and then added together to produce the syntax embedding at position $i$:
\begin{equation*}
    {\rm Emb} (\uscr{x}{i}) = {\rm Emb} (\uscr{p}{i}) + {\rm Emb} (\uscr{l}{i}).
\end{equation*}

\paragraph{Multi-Encoder Attention}

Figure~\ref{fig:syntax-expansion} illustrates the syntax expansion model in \ours.
As shown, the model has two Transformer encoders: a \emph{source encoder} that encodes $\ssrc$, and a \emph{template encoder} that encodes $\stmpl$.
Intuitively, $\stmpl$ regulates $\spred$'s high-level syntactic structure, while the expander fills the details according to $\ssrc$.

How to integrate the information from multiple encoders is critical.
\citet{wang2019task} choose to use a linear layer to combine the encoder outputs and feed the result into the decoder.
The input sequences in their work share the same length, and the tokens at the same position are corresponding to each other, \eg, one input sequence is the sentence and another is its part of speech (POS) tagging.
Our inputs, however, have various lengths, making the simple integration with linear layer infeasible.

Inspired by the multi-head attention mechanism \citep{vaswani2017attention}, we
propose a \emph{multi-encoder attention} mechanism, which extends the concept of multi-head attention by attaching different attention heads to different Transformer encoders (Figure~\ref{fig:syntax-expansion}).
Suppose we have two Transformer encoders with encoding output $H_1\in\R^{m_1
  \times d_{\rm m}}$ and $H_2\in\R^{m_2 \times d_{\rm m}}$, and the decoder's
former layer output $O\in\R^{m_O \times d_{\rm m}}$ where $m_1, m_2$ and $m_O$
are sequence lengths, the multi-encoder attention is calculated as follows:
\begin{gather*}
    C = {\rm Concat} (\uscr{A}{1}_1,\dots, \uscr{A}{h_1}_1, \uscr{A}{1}_2,\dots, \uscr{A}{h_2}_2 ), \\
    \uscr{A}{j}_i = {\rm Attn} (O\cdot \uscr{W}{j}_{Q,i},
    ~H_i\cdot \uscr{W}{j}_{K,i}, ~ H_i\cdot \uscr{W}{j}_{V,i} ),
\end{gather*}
where $\uscr{W}{j}_{Q,i}, \uscr{W}{j}_{K,i} \in \R^{d_{\rm m} \times d_{k}}$, $\uscr{W}{j}_{V,i} \in \R^{d_{\rm m} \times d_{v}}$; $d_{\rm m}$, $d_{k}$ and $d_{v}$ are the vector dimensions; $h_1$ and $h_2$ are the number of decoder heads attached to different encoders.
$\uscr{A}{j}_i \in \R^{m_O\times d_v}$ is the result of the $j$th attention head connected to encoder $i$, calculated in the same way as \citet{vaswani2017attention}:
\begin{equation*}
    {\rm Attn}(Q, K, V) = {\rm Softmax}(\frac{Q\cdot K^{\sf T}}{\sqrt{d_k}})\cdot V.
\end{equation*}
As each matrix $\uscr{A}{j}_i$ has the same dimension, the multi-encoder attention can easily integrate encoder outputs with different sequence lengths through concatenation.
At last, a linear layer is used to fuse the information:
\begin{equation*}
    {\rm Attention}_{\rm MultiEnc} = C \cdot W_O,
\end{equation*}
with projection matrix $W_O\in \R^{(h_1+h_2)d_v \times d_{\rm m}}$.
In this way, multiple Transformer encoders can be attended by a Transformer decoder, even if their encoded sequences have different lengths.

\begin{figure*}[tbp]
    \centerline{\includegraphics[width = 0.8\textwidth]{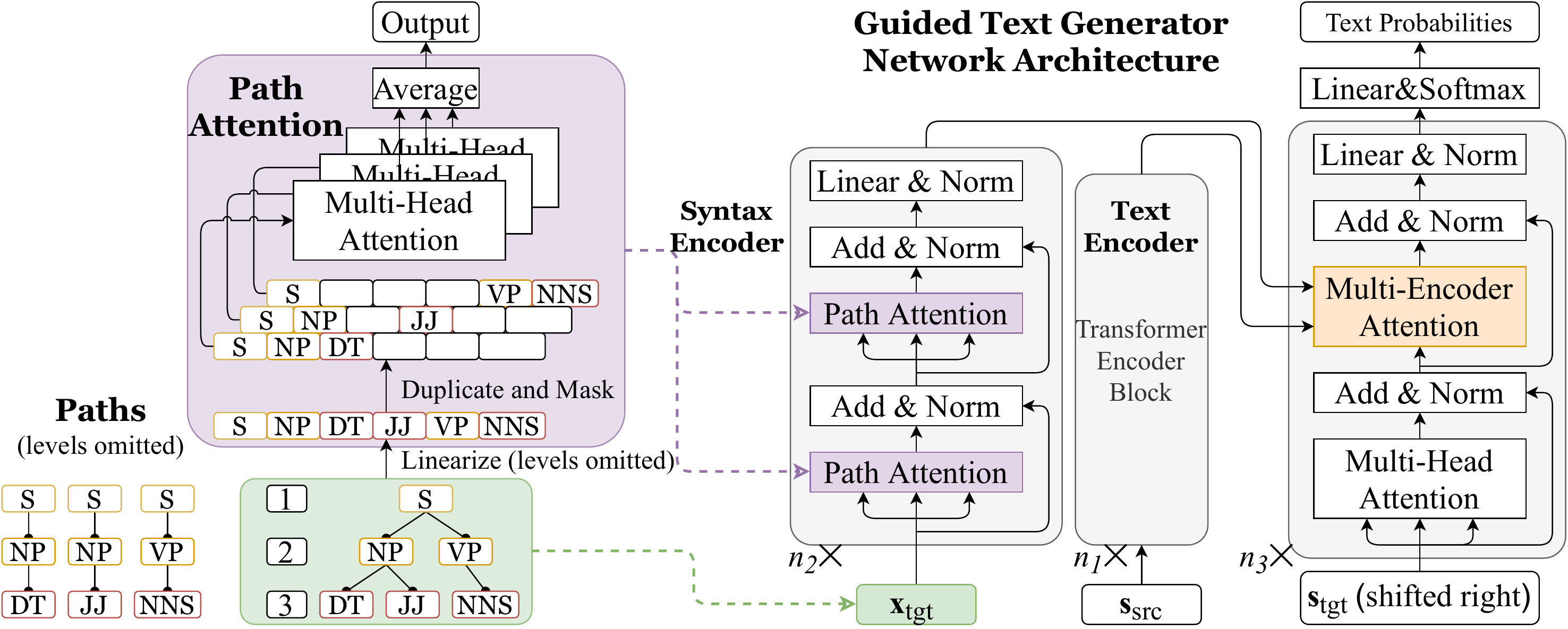}}
    \caption{
    The left figure describes the path attention strategy, and the right figure illustrates the guided generator's network architecture.
    The blanks squares are the masked constituency parse nodes.
    }
    \label{fig:text-generation}
\end{figure*}

\paragraph{Training}
The last decoder block is followed by two classification modules to make two predictions at step $i-1$: the parse node $\uscr{\hat{p}_{\rm tgt}}{i}$ and the level token $\uscr{\hat{l}_{\rm tgt}}{i}$.
Given their probabilities $\uscr{\hat{\y}}{i}_p, \uscr{\hat{\y}}{i}_l$ and the one-hot encoded ground truth $\uscr{\y}{i}_p, \uscr{\y}{i}_l$,
The step loss is the weighted sum of two NLL losses:
\begin{equation*}
    \uscr{{\rm loss}}{i-1}_{\rm syn} = -\alpha \langle \uscr{\log\hat{\y}}{i}_p, \uscr{\y}{i}_p\rangle  -\beta \langle \uscr{\log\hat{\y}}{i}_l, \uscr{\y}{i}_l\rangle,
\end{equation*}
where $\langle \cdot, \cdot \rangle$ is the inner product; $\alpha$ and $\beta$ are loss weights, which are both set to $0.5$ in our work.
The training objective is minimizing the sequence loss, \ie the sum of all step losses.

\subsection{Guided Text Generation}
\label{subsec:text-generation}

The goal of the text generator is to generate the target text $\ttgt$, which is syntactically aligned with the syntactic guidance $\spred$ and meanwhile utilizes the semantics of the source text $\tsrc$.
Similar to the syntax expander, we also use two Transformer encoders---a \emph{syntax encoder} and a \emph{text encoder}---to encode syntax sequence and text sequence separately and a Transformer decoder with the multi-encoder attention mechanism for the generation (see Figure~\ref{fig:text-generation}).
However, as the syntax and text representations belong to different spaces, the situation becomes tricky:
when the provided syntactic structure is specific, chances are particular words are mapped onto the leaf nodes, resulting in a model overfitted to the surface names of syntactic tokens.


\paragraph{Path Attention}
To address the above issue, we propose a \emph{path attention} strategy that forces the network to focus more on general syntactic guidance in higher-level part of the constituency tree.~\footnote{We substitute self-attention layers in the syntax encoder with our path attention layers.}
A path is a route in a tree from the root node to a leaf node (see
Figure~\ref{fig:text-generation}).
Say $O\in\R^{m_O \times d_{\rm m}}$ is the former layer's output, with $m_O$ as
the sequence length and $d_{\rm m}$  as the model dimension.
First, it is duplicated by $n_p$ times ($n_p$ is the total number of paths), forming a set $\{ \uscr{O}{i}, \dots, \uscr{O}{n_p} \}$ in which each element corresponds to a path.
A mask is applied to each element to mask out (set to $-\infty$) those nodes not in the path.
Then, each masked element $\uscr{O_{\rm M}}{i}, i\in [1, n_p]$ is separately fed into the \textit{same} self-attention network:
\begin{gather*}
    C_{i} = {\rm Concat} (\uscr{A}{1}_i,\dots, \uscr{A}{h}_i ) \cdot W_O, \\
    \uscr{A}{j}_i = {\rm Attn} (\uscr{O_{\rm M}}{i}\cdot \uscr{W}{j}_{Q}, \uscr{O_{\rm M}}{i}\cdot \uscr{W}{j}_{K}, \uscr{O_{\rm M}}{i}\cdot \uscr{W}{j}_{V} ),
\end{gather*}
where $W_*^{(*)}$ are learnable weights;
$h$ is the number of attention heads and $j\in[1, h]$.
At last, the results are averaged to form the path attention output:
\begin{equation*}
    {\rm Attention}_{\rm Path} = \frac{1}{n_p}\sum_{i=1}^{n_p}C_{i}.
\end{equation*}

Intuitively, the self-attention mechanism updates each token embedding with a weighted sum of all embeddings in the sequence.
With path attention, however, one node can only exchange information with other nodes that share the same path.
To acquire information from a node outside its path, one must turn to their
common ancestor, who is able to get the desired information from former path attention layers, forcing the ancestors (higher-level guidance) to be more heavily attended to than the descendants.
The path attention is executed twice in each block so that the information carried by each node flows around the entire sequence.

The reason we do not include the path attention strategy in the syntax expander is that the input and output of that model are both linearized parse trees.
Using path attention in the encoder would create a mismatch between the encoding and decoding process that harms model performance.

\paragraph{Training}
The guided generator is trained by minimizing the NLL loss between the probability $\uscr{\hat{\y}}{i}_s$ and the one-hot encoded ground truth word $\uscr{\y}{i}_s$:
\begin{equation*}
    {\rm loss}_{\rm txt} = -\sum_{i=1}^{M} \langle \log\uscr{\hat{\y}}{i}_s, \uscr{\y}{i}_s\rangle,
\end{equation*}
where $M$ is the sequence length.
During inference, the syntax guidance of the text generator can be either the full-fledged target parse tree $\stgt$ or the output of the syntax expander $\spred$.

%% file: 4-experiments.tex
\begin{table*}[tbp]\small
    \centering
    \begin{tabular}{c|ccccc|cc}
    \toprule
    Model & BLEU $\uparrow$ & ROUGE-1 $\uparrow$ & ROUGE-2 $\uparrow$ & ROUGE-L $\uparrow$ & METEOR $\uparrow$ & TED-f $\downarrow$ & TED-8 $\downarrow$\\
    \midrule
    VGVAE & 13.6 & 44.7 & 21.0 & 48.3 & 24.8 & 6.7 & - \\
    SCPN & 23.23 & 53.21 & 31.05 & 57.22 & 51.91 & 6.55 & 6.21 \\
    Transformer & 46.00 & 73.32 & 54.45 & 75.47 & 73.50 & 6.33 & 5.95 \\
    \midrule
    w/o syn & 15.96 & 50.11 & 23.83 & 49.68 & 46.84 & 11.88 & 11.44 \\
    w/o txt & 13.41 & 39.74 & 20.62 & 44.72 & 37.40 & 6.35 & 5.89 \\
    w/o PA & 38.91 & 68.01 & 47.78 & 70.67 & 67.26 & 6.36 & \textbf{5.88} \\
    \guigen & \textbf{48.03} & \textbf{74.53} & \textbf{56.05} & \textbf{76.65} & \textbf{75.02} & \textbf{6.23} & 5.89 \\
    \bottomrule
    \end{tabular}
    \caption{Text generation results with the ground truth target parse $\stgt$ as syntactic guidance.
    ``TG'' represents the text generator.
    ``Transformer'' is introduced in \cref{subsect:setup}.
    ``w/o syn'' is a Transformer without syntactic constraint whereas ``w/o txt'' has no source text input.
    ``w/o PA'' is the \guigen without path attention strategy applied to the syntax encoder.
    The arrows show the direction where better performance is.
    }
    \label{tb:full_guidance_results}
\end{table*}

\section{Experiments}
\label{sec:experiments}

\subsection{Setup}
\label{subsect:setup}

\paragraph{Task and Data Preperation}

We evaluate \ours on the paraphrase generation task.
Following \citet{iyyer2018adversarial}, we first evaluate the guided text generator's ability to follow the syntactic guidance by predicting paraphrases with the source text $\tsrc$ and the target parse $\stgt$.
Then, we assess the performance of our syntax expander by predicting text using the constituency tree $\spred$ expanded from the template parse $\stmpl$.

Our dataset is a subset of ParaNMT-50M \citep{Wieting_2018} provided by \citet{chen2019controllable}.
In our work, the number of total text sub-word tokens is $16,000$.
The constituency parsing tool is provided by AllenNLP \citep{Gardner_2018}.
The sentences whose parse trees contain infrequent tokens are excluded, and all trees are truncated to $8$-level for simplicity, leaving $74$ parse nodes and $12$ level tokens in total.
The dataset is standardized by removing all paraphrase pairs whose text or syntax sequences are longer than $50$ or with non-ASCII characters.
After the pre-processing, $447,536$ paraphrase pairs remain in the dataset, in which $90\%$ are randomly selected for training and the rest for validation.
Independent from them, $500$ and $800$ high-quality paraphrase pairs manually annotated by \citet{chen2019controllable} are used for model development and evaluation.
The training details of \ours are described in the Appendix.

\paragraph{Baselines}

We include three baselines:
\begin{itemize}
    \item SCPN \citep{iyyer2018adversarial} is based on LSTM, attention and copy mechanism \citep{see2017get}. It is trained and evaluated on the same dataset as ours, except that their parse tree is linearized to the bracketed format. The network hyper-parameters are set to default.
    \item VGVAE \citep{chen2019controllable} uses reference sentences as the
      syntactic constraint instead of constituency trees. 
    \item A standard Transformer \citep{vaswani2017attention} for syntax-guided text generation.
    We concatenate the input syntax guidance with the source text and feed the connected sequence into the model to generate target text.
\end{itemize}
In addition to the above baselines, we also include ablations of our model to study the effectiveness of different components in \ours.


\subsection{Quantitative Evaluation}
\label{subsect:quantity}
To evaluate the performance of different methods, we
use three metrics for semantic congruity and one for syntactic similarity.
The semantic metrics are: 1) BLEU \citep{papineni2002bleu}; 2) ROUGE \citep{lin2004rouge}, including ROUGE-1, ROUGE-2, ROUGE-L; and 3) METEOR \citep{banerjee2005meteor}~\footnote{BLEU \& METEOR: \url{https://www.nltk.org/}; ROUGE: \url{https://github.com/Diego999/py-rouge}.}.
To assess syntactic alignment, we calculate the tree edit distance (TED)~\footnote{\url{https://github.com/JoaoFelipe/apted}} between the generated and target sentences' constituency parse trees.
It measures the number of insertion, rotation and removal operations needed for changing one tree to another.

\begin{table*}[tbp]\small
    \centering
    \begin{tabular}{cc|ccccc|c}
    \toprule
    SE & TG & BLEU $\uparrow$ & ROUGE-1 $\uparrow$ & ROUGE-2 $\uparrow$ & ROUGE-L $\uparrow$ & METEOR $\uparrow$ & N-TED-f $\downarrow$\\
    \midrule
    SCPN & SCPN & 11.83 & 41.83 & 20.36 & 45.63 & 38.59 & 0.5074 \\
    SCPN & \guigen & 19.52 & 55.14 & 31.11 & 57.89 & 51.69 & 0.5125 \\
    \midrule
    w/ PA &  \guigen & 19.47 & 53.06 & 29.39 & 55.87 & 50.81 & 0.4946 \\
    w/o tmpl & \guigen & 13.29 & 45.56 & 18.62 & 45.38 & 42.48 & 0.6843\\
    $\stmpl$ & \guigen & 20.22 & 56.00 & 32.73 & 58.57 & 50.63 & 0.6227 \\
    \synexpan & \guigen & \textbf{26.27} & \textbf{61.10} & \textbf{37.13} & \textbf{63.04} & \textbf{59.88} & \textbf{0.4732} \\
    \bottomrule
    \end{tabular}
    \caption{Synthetic evaluation of the syntax expansion and text generation models.
    SE and TG are syntax expansion and text generation models respectively.
    ``w/o tmpl'' uses only source parse to predict target parse.
    $\stmpl$ indicates that the template parses are directly fed into the text generation model without expansion.
    ``w/ PA'' is our syntax expander with path attention applied to its source syntax encoder.
    }
    \label{tb:tmpl_guidance_results}
\end{table*}

\paragraph{Text Generation with Target Parse}

When the full constituency trees of target sentences $\stgt$ are given as syntactic guidance, Table~\ref{tb:full_guidance_results} shows that our generator has better semantic and syntactic performance than SCPN and VGVAE by doubling their BLEU scores as well as presenting a smaller TED.
Comparing our generator with the standard Transformer, we find that encoding different information separately is a better way than mixing them together in the same encoder.

The table also shows that both source text and syntactic guidance are indispensable.
Paraphrasing with only source text gives fair semantics, but completely fails to control the syntactic structure;
whereas text generated without source text, as one may predict, has fair syntactic structure but poor semantics.
The last two rows indicate that path attention significantly contributes to the semantic expression without losing the syntactic integrity.
The results support our claim that it encourages the model to attend to higher-level guidance and learn the real syntactic structure instead of a parse-to-word mapping.

\begin{figure}[tbp]
    \centering {
        \subfloat[TED (the lower the better).] {
            \label{subfig:ted}
            \includegraphics[width=0.48\textwidth]{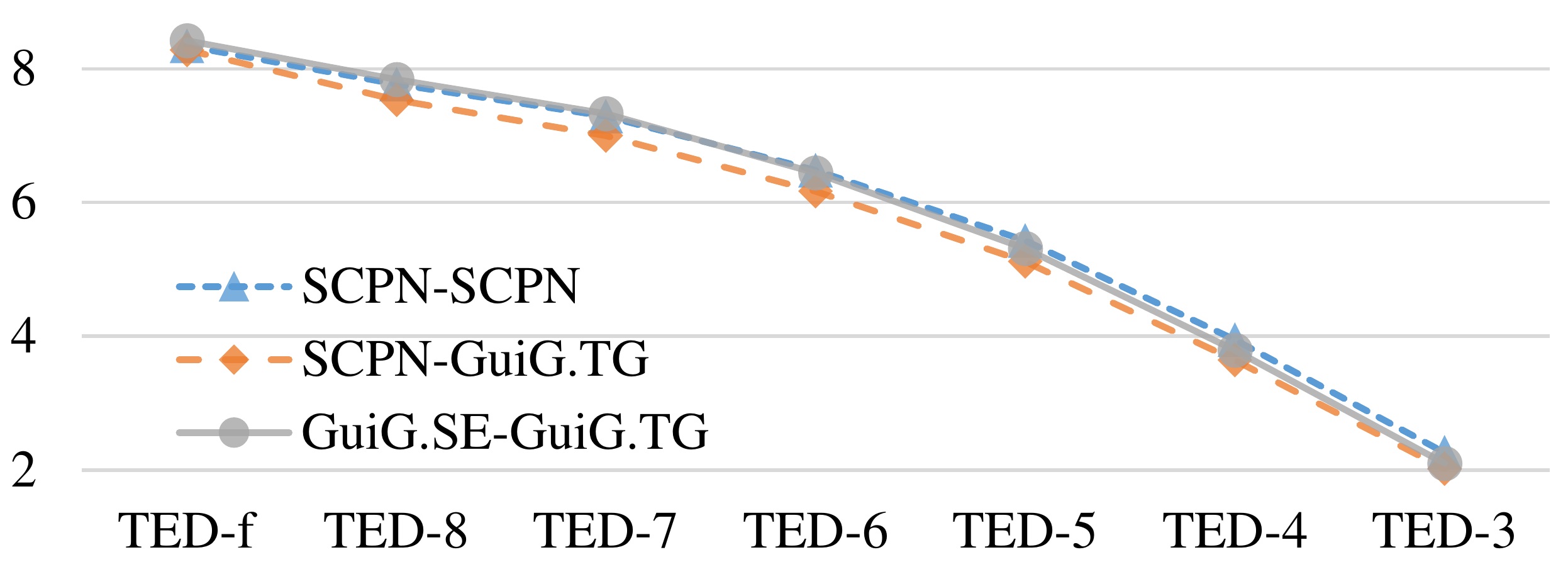}
        }
        \hfil
        \subfloat[Normalized TED (the lower the better).] {
            \label{subfig:n-ted}
            \includegraphics[width=0.48\textwidth]{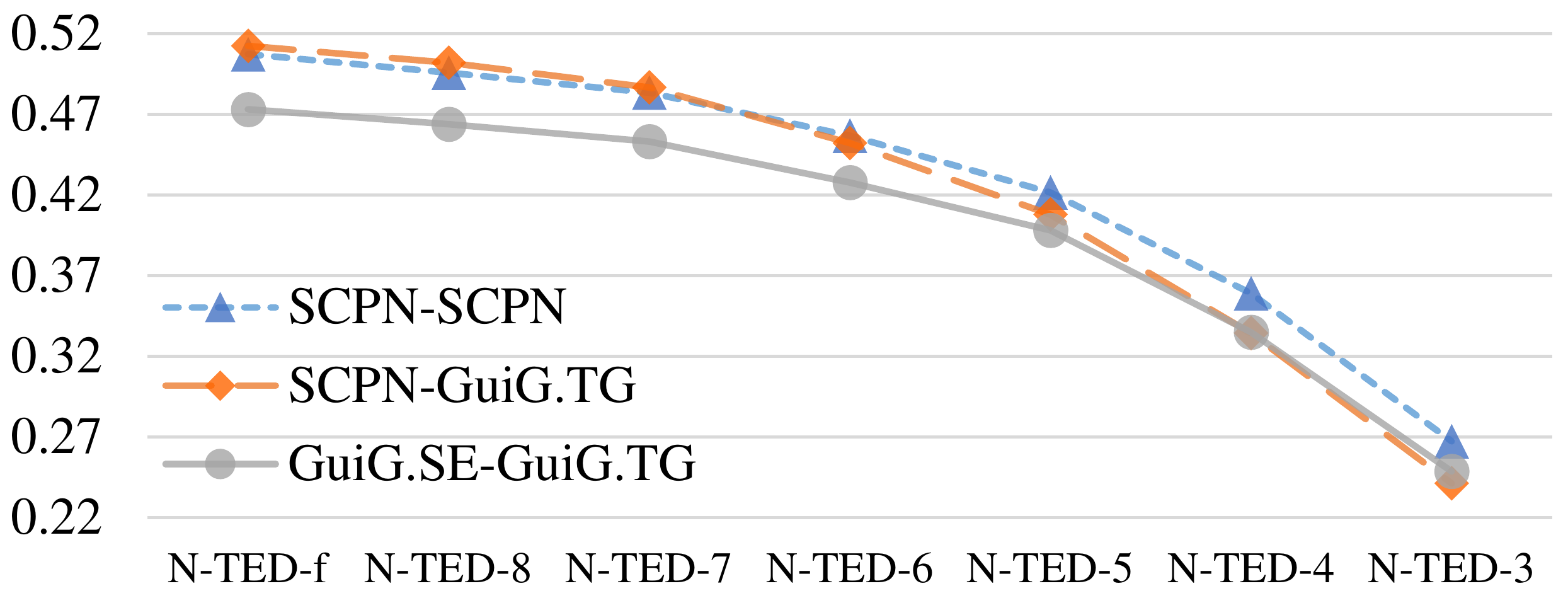}
        }
    }
    \caption{Original and normalized tree edit distances between target and generated sentences.
    }
    \label{fig:tmpl-guidance-results}
\end{figure}

\paragraph{Text Generation with Expanded Parse}

Table~\ref{tb:tmpl_guidance_results} and Figure~\ref{fig:tmpl-guidance-results} present the performance of text generators when their syntax guidance comes from SCPN's syntax expansion model.
With the same expanded parse, our text generator again demonstrates better semantic results and similar syntactic results.
This proves our text generator's superiority is independent of the source of syntactic guidance.

\paragraph{Syntax Expansion}

Since our ultimate goal is to generate text, we indirectly evaluate the syntax expander through the assessment of the text generated under the guidance of the expanded parse $\spred$.
To make it fair, here we uniformly use our text generator to generate sentences with $\spred$ expanded from different syntax expanders.
Also, the maximum syntax sequence length of SCPN is set to $150$ as their linearization method takes $3$ times the length as ours.
In addition, TED unfairly favors shorter generated sentences.
Therefore we report its normalized version---N-TED, \ie, TED divided by the number of nodes in a tree, when the expanded syntax sequences are of different lengths.
Specifically, we report N-TED-$\ell$ and ``N-TED-f'' to give a full description of how the consistency trees of generated sentences aligned to the target syntax at different levels.
$\ell$ indicates how many levels of parse trees are kept when we compare the syntax of the generated and target sentences, and ``f'' (full) means the parse trees are intact.

Table~\ref{tb:tmpl_guidance_results} mainly presents the semantic results whereas Figure~\ref{fig:tmpl-guidance-results} illustrates the detailed syntactic performance.
Comparing our syntax expander to SCPN's, one can see that our model is more capable of predicting reasonable syntactic structures, with which the text generator can generate sentences more semantically analogous to the targets.
Although Figure~\ref{subfig:ted} implies that SCPN's syntax expansion model produces better syntactic results, it is because our syntax expander predicts larger parse trees.
When the trees are larger, the text generator is prone to output longer sentences, biasing the evaluation against our model.
Removing the interference of length, our syntax expander gives better scores when the trimmed syntax trees are deeper than $4$ levels, as shown in Figure~\ref{subfig:n-ted}.
Benefitted from the copy mechanism, SCPN is more capable of maintaining the template parse but disadvantaged in giving convincing predictions.

The results in the table also testify the disadvantage of using path attention in the syntax expander, as discussed in \cref{subsec:text-generation}.
Expanding the template parse tree $\stmpl$ without the source parse tree $\ssrc$ is impractical since the result would lose the ability to fit the source syntax;
and generating target parse solely based on the source parse fails to properly control the syntactic structure due to the absence of template parse.
If we guide the text generator directly with the template parse $\stmpl$, the model is not able to acquire adequate syntactic information and presents poor results.

\begin{table}[tbp]\small
    \centering
    \begin{tabular}{c|cc|cc}
    \toprule
    \multirow{2}{0.3in}{\centering syntax input}
     & \multicolumn{2}{c|}{MODEL} & \multicolumn{2}{c}{SCORES} \\
     & SE & TG & Semantic & Syntactic \\
    \midrule
    \multirow{2}{0.3in}{\centering $\stgt$}
    & - & SCPN & 2.93 & 4.00  \\
    & - & \guigen & \textbf{4.21} & \textbf{4.67}  \\
    \midrule
    \multirow{3}{0.3in}{\centering $\stmpl$}
    & SCPN & SCPN & 2.71 & 3.48 \\
    & SCPN & \guigen & 3.52 & 3.88 \\
    & \synexpan & \guigen & \textbf{3.84} & \textbf{4.10} \\
    \bottomrule
    \end{tabular}
    \caption{
    Human evaluation scores.
    $\stgt$ is the case where we use target parses $\stgt$ as generator's syntactic guidance without expansion model.
    $\stmpl$ is the case where the generator takes the expanded parse.
    }
    \label{tb:human-evaluation}
\end{table}

\subsection{Human Evaluation}

We perform a crowdsourced evaluation of the semantics and syntax of the generated sentences.
$200$ examples are randomly selected.
Each of them is evaluated by three workers in the way of scoring the semantic and syntactic similarities between the generated and target sentences from $1$ to $5$, the higher the better.

The results presented in Table~\ref{tb:human-evaluation} are largely parallel to the objective metrics (\cref{subsect:quantity}).
Compared with SCPN, our text generator generates more semantically reasonable text with the syntactic guidance comes from either the target parse $\stgt$ ($1$st and $2$nd rows) or the parse $\spred$ expanded from template $\stmpl$ by SCPN ($3$rd and $4$th rows).
The $4$th and $5$th rows in the table prove that our syntax expander also contributes to the performance improvement.

Analyzing the sentences that get low ($\leqslant 2$) semantic or syntactic average scores, we find our text generator sometimes suffers from several defects:
1) the generated sentence is
semantically opposite to the target, especially when the source text has multi-negation;
2) one word gets repeated for several times; and
3) incomprehensible words are given due to the usage of BPE.
These issues are universal in all text generation models and deserve further investigation.
However, these situations are rare, and our method generates fluent and well-structured sentences most of the time.

\subsection{Qualitative Analysis}
\label{subsect:quality}

\begin{table}[tbp]\small
    \centering
    \begin{tabular}{p{2.8in}}
    \toprule
    \textbf{src:} you told me to look for the wolf where his prey can be found. \\
    \textbf{tgt:} you said i would find a wolf where i can find its prey. \\
    \textbf{SCPN:} i thought you'd find it if the wolf saw the wolf. \\
    \textbf{\ours:} you said i should look for a wolf where he could find his prey. \\
    \midrule
    \textbf{src:} wounds on the body may easily be healed, but emotional scars do not go away so easily. \\
    \textbf{tgt:} physical injuries will heal, but it is not so easy for scars on the soul. \\
    \textbf{SCPN:} it was a lot of the body, but he 'll have no way of the body. \\
    \textbf{\ours:} the wounds on the body can be healed easily, but emotional scars do n't go so easily. \\
    \midrule
    \textbf{src:} we need to further strengthen the agency's capacities. \\
    \textbf{tgt:} the capacity of this office needs to be reinforced even further. \\
    \textbf{SCPN:} the possibility of the agency is to survive. \\
    \textbf{\ours:} the capacity of the agency needs to be further strengthened. \\
    \bottomrule
    \end{tabular}
    \caption{Examples generated with expanded parse $\spred$.
    }
    \label{tb:tg_ep}
\end{table}

\paragraph{Text Generation with Expanded Parse}

Table~\ref{tb:tg_ep} shows a few examples generated under the guidance of expanded parses.
It can be observed that most of the time the semantic meanings of source sentences are well-preserved while the syntactic structures are successfully transferred.
However, in some cases, the predicted text fails to entirely follow the references' syntactic structures due to the imperfection of the syntax expansion model.
In the second example, the syntactic distinction of the source and reference sentences lies in micro rather than macro scale.
Consequently, the predicted parse copies heavily from the source parse, making the generated sentence more similar to the source text instead of the target.
Nonetheless, compared with SCPN, our model is more capable of using appropriate expression and suffers less from the repeated words issue, leading to more comprehensible and better-structured sentences.

\begin{table}[tbp]\small
    \centering
    \begin{tabular}{p{2.8in}}
    \toprule
    \textbf{PARSE NODE:} S NP PRP VP MD VP .\\
    \textbf{LEVEL:} 1 2 3 2 3 3 2\\
    \midrule
    \textbf{src:} he believed his son had died in a terrorist attack. \\
    \textbf{gen:} he would believe his son was killed in a terrorist attack.\\
    \midrule
    \textbf{src:} she seems to have written a book about driving. \\
    \textbf{gen:} she must have written a book on the driver.\\
    \midrule
    \textbf{src:} it is hard for me to imagine where they could be hiding it underground.\\
    \textbf{gen:} i'd be difficult to imagine where they 're hiding him underground .\\
    \midrule
    \midrule
    \textbf{PARSE NODE:} S NP PRP VP VBZ NP .\\
    \textbf{LEVEL:} 1 2 3 2 3 3 2\\
    \midrule
    \textbf{src:} there were 50 bucks' worth of merchandise stolen by a fucker today . \\
    \textbf{gen:} it's 50 bucks for the kind of thing stolen by a motherfucker.\\
    \midrule
    \textbf{src:} there's an intelligent way to approach marriage. \\
    \textbf{gen:} it is a smart way to approach the wedding.\\
    \midrule
    \textbf{src:} stealing state secrets was one thing he was framed for by frank. \\
    \textbf{gen:} it's a part of the theft of state secrets that frank has been framed.\\
    \bottomrule
    \end{tabular}
    \caption{Generated examples with frequently appeared template constituency parse trees.}
    \label{tb:tg_cp}
\end{table}

\paragraph{Text Generation with Common Templates}

We take a step further and demonstrate our model's ability to generate sentences from the templates that appear most frequently in the dataset.
Table~\ref{tb:tg_cp} shows that the sentences generated with the same template parses have similar high-level structures.
Moreover, the semantic analogy between the source and generated sentences proves our method's ability to successfully keep the semantics during the syntax transfer process.

%% file: 5-related-works.tex
\section{Related Works}

Constrained text generation has attracted much attention in recent years.
Categorized by the object to be controlled, there are two tracks of works: one seeks to manipulate the semantic attributes \citep{hu2017toward, li2018delete, li2018guiding, Yin-etal-2019-utilizing, Wang-etal-2019-controllable}.
For example, \citet{hu2017toward} generate text with specified sentiments, whereas \citet{li2018delete} and \citet{Wang-etal-2019-controllable} try to transfer the sentiments or styles of the source sentences.
The other track, to which our research belongs, focuses on making generated text follow a particular style or structure \citep{Niu2017ASO, ficler2017controlling, fu2018style, liu2018controlling, iyyer2018adversarial, chen2019controllable, li-etal-2019-decomposable, balasubramanian2020polarizedvae}.
For instance, \citet{Niu2017ASO} constrain the output styles in neural machine translation task and \citet{liu2018controlling} impose length limitation to the summarization.

Based on the constraint source, syntactically controlled text generation models can be further divided into three groups.
The first group \citep{chen2019multi, bao2019generating, balasubramanian2020polarizedvae} takes sentences as syntactic exemplars.
They attempt to disentangle the semantic and syntactic representations into different VAE \citep{kingma2013autoencoding} latent spaces during training, and then use the exemplar to assign a prior distribution to the syntactic latent space at the inference stage.
The second group \citep{iyyer2018adversarial, zhang2019syntax} directly employs the constituency tree as an auxiliary input, controlling the syntax of generated text with the structure specified by it.
Instead of importing externally, the third group \citep{wiseman2018learning, akoury-etal-2019-syntactically, casas2020syntaxdriven} learns the syntax guidance from the training data and apply it in the generation phrase in return.

Considering that the fully specified exemplar sentences are hard to be effectively retrieved \citep{goyal2020neural}, we follow \citet{iyyer2018adversarial} and use constituency trees as the syntax guidance.
We further take advantage of the parallel attribute of Transformer \citep{vaswani2017attention} to accommodate the tree structure in the encoding process.
There are works \citep{eriguchi-etal-2016-tree, chen-etal-2017-improved, ding2019recurrent} that adapt the recurrent encoder to the trees, but the transition matrix that RNNs depend on is less effective than our attention mechanism, especially when the tree is large.

%% file: 6-conclusion.tex
\section{Conclusion}

We have proposed a novel syntactically guided text generation method \ours.~\footnote{The code and data are available at \url{https://github.com/Yinghao-Li/GuiGen}.}
It expands the template constituency parse tree to a full-fledged tree, using it as the syntactic constraint to guide the text generation process.
A syntax expander based on the multi-encoder Transformer is designed to predict a convincing target parse tailored for the source text, and a guided text generator powered by path attention strategy is introduced to generate text that has the semantics specified by the source text as well as complies with the syntactic guidance.
Evaluated on the paraphrasing task, ablation studies justify the necessity of the components of our method, while quantitative and qualitative experiments demonstrate our method's ability to generates more semantically reasonable and syntactically aligned sentences than SOTA baselines.
We believe our method can play an important role in style transfer and text data augmentation applications.


%% file: main.bbl
\begin{thebibliography}{38}
\expandafter\ifx\csname natexlab\endcsname\relax\def\natexlab#1{#1}\fi

\bibitem[{Akoury et~al.(2019)Akoury, Krishna, and
  Iyyer}]{akoury-etal-2019-syntactically}
Nader Akoury, Kalpesh Krishna, and Mohit Iyyer. 2019.
\newblock \href {https://doi.org/10.18653/v1/P19-1122} {Syntactically
  supervised transformers for faster neural machine translation}.
\newblock In \emph{Proceedings of the 57th Annual Meeting of the Association
  for Computational Linguistics}, pages 1269--1281, Florence, Italy.
  Association for Computational Linguistics.

\bibitem[{Bahdanau et~al.(2014)Bahdanau, Cho, and Bengio}]{bahdanau2014neural}
Dzmitry Bahdanau, Kyunghyun Cho, and Yoshua Bengio. 2014.
\newblock \href {http://arxiv.org/abs/1409.0473} {Neural machine translation by
  jointly learning to align and translate}.
\newblock Cite arxiv:1409.0473 Comment: Accepted at ICLR 2015 as oral
  presentation.

\bibitem[{Balasubramanian et~al.(2020)Balasubramanian, Kobyzev, Bahuleyan,
  Shapiro, and Vechtomova}]{balasubramanian2020polarizedvae}
Vikash Balasubramanian, Ivan Kobyzev, Hareesh Bahuleyan, Ilya Shapiro, and Olga
  Vechtomova. 2020.
\newblock \href {http://arxiv.org/abs/2004.10809} {Polarized-vae: Proximity
  based disentangled representation learning for text generation}.

\bibitem[{Banerjee and Lavie(2005)}]{banerjee2005meteor}
Satanjeev Banerjee and Alon Lavie. 2005.
\newblock \href {https://www.aclweb.org/anthology/W05-0909} {{METEOR}: An
  automatic metric for {MT} evaluation with improved correlation with human
  judgments}.
\newblock In \emph{Proceedings of the {ACL} Workshop on Intrinsic and Extrinsic
  Evaluation Measures for Machine Translation and/or Summarization}, pages
  65--72, Ann Arbor, Michigan. Association for Computational Linguistics.

\bibitem[{Bao et~al.(2019)Bao, Zhou, Huang, Li, Mou, Vechtomova, Dai, and
  Chen}]{bao2019generating}
Yu~Bao, Hao Zhou, Shujian Huang, Lei Li, Lili Mou, Olga Vechtomova, Xin-yu Dai,
  and Jiajun Chen. 2019.
\newblock \href {https://doi.org/10.18653/v1/P19-1602} {Generating sentences
  from disentangled syntactic and semantic spaces}.
\newblock In \emph{Proceedings of the 57th Annual Meeting of the Association
  for Computational Linguistics}, pages 6008--6019, Florence, Italy.
  Association for Computational Linguistics.

\bibitem[{Casas et~al.(2020)Casas, Fonollosa, and
  Costa-jussà}]{casas2020syntaxdriven}
Noe Casas, José A.~R. Fonollosa, and Marta~R. Costa-jussà. 2020.
\newblock \href {http://arxiv.org/abs/2004.02211} {Syntax-driven iterative
  expansion language models for controllable text generation}.

\bibitem[{Chen et~al.(2017)Chen, Huang, Chiang, and
  Chen}]{chen-etal-2017-improved}
Huadong Chen, Shujian Huang, David Chiang, and Jiajun Chen. 2017.
\newblock \href {https://doi.org/10.18653/v1/P17-1177} {Improved neural machine
  translation with a syntax-aware encoder and decoder}.
\newblock In \emph{Proceedings of the 55th Annual Meeting of the Association
  for Computational Linguistics (Volume 1: Long Papers)}, pages 1936--1945,
  Vancouver, Canada. Association for Computational Linguistics.

\bibitem[{Chen et~al.(2019{\natexlab{a}})Chen, Tang, Wiseman, and
  Gimpel}]{chen2019controllable}
Mingda Chen, Qingming Tang, Sam Wiseman, and Kevin Gimpel. 2019{\natexlab{a}}.
\newblock \href {https://doi.org/10.18653/v1/P19-1599} {Controllable paraphrase
  generation with a syntactic exemplar}.
\newblock In \emph{Proceedings of the 57th Annual Meeting of the Association
  for Computational Linguistics}, pages 5972--5984, Florence, Italy.
  Association for Computational Linguistics.

\bibitem[{Chen et~al.(2019{\natexlab{b}})Chen, Tang, Wiseman, and
  Gimpel}]{chen2019multi}
Mingda Chen, Qingming Tang, Sam Wiseman, and Kevin Gimpel. 2019{\natexlab{b}}.
\newblock \href {https://doi.org/10.18653/v1/N19-1254} {A multi-task approach
  for disentangling syntax and semantics in sentence representations}.
\newblock In \emph{Proceedings of the 2019 Conference of the North {A}merican
  Chapter of the Association for Computational Linguistics: Human Language
  Technologies, Volume 1 (Long and Short Papers)}, pages 2453--2464,
  Minneapolis, Minnesota. Association for Computational Linguistics.

\bibitem[{Ding and Tao(2019)}]{ding2019recurrent}
Liang Ding and Dacheng Tao. 2019.
\newblock \href {http://arxiv.org/abs/1908.06559} {Recurrent graph syntax
  encoder for neural machine translation}.

\bibitem[{Eriguchi et~al.(2016)Eriguchi, Hashimoto, and
  Tsuruoka}]{eriguchi-etal-2016-tree}
Akiko Eriguchi, Kazuma Hashimoto, and Yoshimasa Tsuruoka. 2016.
\newblock \href {https://doi.org/10.18653/v1/P16-1078} {Tree-to-sequence
  attentional neural machine translation}.
\newblock In \emph{Proceedings of the 54th Annual Meeting of the Association
  for Computational Linguistics (Volume 1: Long Papers)}, pages 823--833,
  Berlin, Germany. Association for Computational Linguistics.

\bibitem[{Ficler and Goldberg(2017)}]{ficler2017controlling}
Jessica Ficler and Yoav Goldberg. 2017.
\newblock \href {https://doi.org/10.18653/v1/W17-4912} {Controlling linguistic
  style aspects in neural language generation}.
\newblock In \emph{Proceedings of the Workshop on Stylistic Variation}, pages
  94--104, Copenhagen, Denmark. Association for Computational Linguistics.

\bibitem[{Fu et~al.(2018)Fu, Tan, Peng, Zhao, and Yan}]{fu2018style}
Zhenxin Fu, Xiaoye Tan, Nanyun Peng, Dongyan Zhao, and Rui Yan. 2018.
\newblock Style transfer in text: Exploration and evaluation.
\newblock In \emph{Thirty-Second AAAI Conference on Artificial Intelligence}.

\bibitem[{Gardner et~al.(2018)Gardner, Grus, Neumann, Tafjord, Dasigi, Liu,
  Peters, Schmitz, and Zettlemoyer}]{Gardner_2018}
Matt Gardner, Joel Grus, Mark Neumann, Oyvind Tafjord, Pradeep Dasigi,
  Nelson~F. Liu, Matthew Peters, Michael Schmitz, and Luke Zettlemoyer. 2018.
\newblock \href {https://doi.org/10.18653/v1/W18-2501} {{A}llen{NLP}: A deep
  semantic natural language processing platform}.
\newblock In \emph{Proceedings of Workshop for {NLP} Open Source Software
  ({NLP}-{OSS})}, pages 1--6, Melbourne, Australia. Association for
  Computational Linguistics.

\bibitem[{Goyal and Durrett(2020)}]{goyal2020neural}
Tanya Goyal and Greg Durrett. 2020.
\newblock \href {http://arxiv.org/abs/2005.02013} {Neural syntactic preordering
  for controlled paraphrase generation}.

\bibitem[{Hochreiter and Schmidhuber(1997)}]{hochreiter1997long}
Sepp Hochreiter and J\"{u}rgen Schmidhuber. 1997.
\newblock \href {https://doi.org/10.1162/neco.1997.9.8.1735} {Long short-term
  memory}.
\newblock \emph{Neural Comput.}, 9(8):1735--1780.

\bibitem[{Hu et~al.(2017)Hu, Yang, Liang, Salakhutdinov, and
  Xing}]{hu2017toward}
Zhiting Hu, Zichao Yang, Xiaodan Liang, Ruslan Salakhutdinov, and Eric~P Xing.
  2017.
\newblock Toward controlled generation of text.
\newblock In \emph{Proceedings of the 34th International Conference on Machine
  Learning-Volume 70}, pages 1587--1596. JMLR. org.

\bibitem[{Iyyer et~al.(2018)Iyyer, Wieting, Gimpel, and
  Zettlemoyer}]{iyyer2018adversarial}
Mohit Iyyer, John Wieting, Kevin Gimpel, and Luke Zettlemoyer. 2018.
\newblock \href {https://doi.org/10.18653/v1/N18-1170} {Adversarial example
  generation with syntactically controlled paraphrase networks}.
\newblock In \emph{Proceedings of the 2018 Conference of the North {A}merican
  Chapter of the Association for Computational Linguistics: Human Language
  Technologies, Volume 1 (Long Papers)}, pages 1875--1885, New Orleans,
  Louisiana. Association for Computational Linguistics.

\bibitem[{Kingma and Welling(2014)}]{kingma2013autoencoding}
Diederik~P. Kingma and Max Welling. 2014.
\newblock \href {http://arxiv.org/abs/1312.6114} {Auto-encoding variational
  bayes}.
\newblock In \emph{2nd International Conference on Learning Representations,
  {ICLR} 2014, Banff, AB, Canada, April 14-16, 2014, Conference Track
  Proceedings}.

\bibitem[{Li et~al.(2018{\natexlab{a}})Li, Xu, Li, and Gao}]{li2018guiding}
Chenliang Li, Weiran Xu, Si~Li, and Sheng Gao. 2018{\natexlab{a}}.
\newblock \href {https://doi.org/10.18653/v1/N18-2009} {Guiding generation for
  abstractive text summarization based on key information guide network}.
\newblock In \emph{Proceedings of the 2018 Conference of the North {A}merican
  Chapter of the Association for Computational Linguistics: Human Language
  Technologies, Volume 2 (Short Papers)}, pages 55--60, New Orleans, Louisiana.
  Association for Computational Linguistics.

\bibitem[{Li et~al.(2018{\natexlab{b}})Li, Jia, He, and Liang}]{li2018delete}
Juncen Li, Robin Jia, He~He, and Percy Liang. 2018{\natexlab{b}}.
\newblock \href {https://doi.org/10.18653/v1/N18-1169} {Delete, retrieve,
  generate: a simple approach to sentiment and style transfer}.
\newblock In \emph{Proceedings of the 2018 Conference of the North {A}merican
  Chapter of the Association for Computational Linguistics: Human Language
  Technologies, Volume 1 (Long Papers)}, pages 1865--1874, New Orleans,
  Louisiana. Association for Computational Linguistics.

\bibitem[{Li et~al.(2019{\natexlab{a}})Li, Qiu, Tang, Chen, Zhao, and
  Yan}]{li-etal-2019-Insufficient}
Juntao Li, Lisong Qiu, Bo~Tang, Dongmin Chen, Dongyan Zhao, and Rui Yan.
  2019{\natexlab{a}}.
\newblock \href {https://doi.org/10.1609/aaai.v33i01.33016698} {Insufficient
  data can also rock! learning to converse using smaller data with
  augmentation}.
\newblock In \emph{The Thirty-Third {AAAI} Conference on Artificial
  Intelligence, {AAAI} 2019, The Thirty-First Innovative Applications of
  Artificial Intelligence Conference, {IAAI} 2019, The Ninth {AAAI} Symposium
  on Educational Advances in Artificial Intelligence, {EAAI} 2019, Honolulu,
  Hawaii, USA, January 27 - February 1, 2019}, pages 6698--6705. {AAAI} Press.

\bibitem[{Li et~al.(2019{\natexlab{b}})Li, Jiang, Shang, and
  Liu}]{li-etal-2019-decomposable}
Zichao Li, Xin Jiang, Lifeng Shang, and Qun Liu. 2019{\natexlab{b}}.
\newblock \href {https://doi.org/10.18653/v1/P19-1332} {Decomposable neural
  paraphrase generation}.
\newblock In \emph{Proceedings of the 57th Annual Meeting of the Association
  for Computational Linguistics}, pages 3403--3414, Florence, Italy.
  Association for Computational Linguistics.

\bibitem[{Lin(2004)}]{lin2004rouge}
Chin-Yew Lin. 2004.
\newblock \href {https://www.aclweb.org/anthology/W04-1013} {{ROUGE}: A package
  for automatic evaluation of summaries}.
\newblock In \emph{Text Summarization Branches Out}, pages 74--81, Barcelona,
  Spain. Association for Computational Linguistics.

\bibitem[{Liu et~al.(2018)Liu, Luo, and Zhu}]{liu2018controlling}
Yizhu Liu, Zhiyi Luo, and Kenny Zhu. 2018.
\newblock \href {https://doi.org/10.18653/v1/D18-1444} {Controlling length in
  abstractive summarization using a convolutional neural network}.
\newblock In \emph{Proceedings of the 2018 Conference on Empirical Methods in
  Natural Language Processing}, pages 4110--4119, Brussels, Belgium.
  Association for Computational Linguistics.

\bibitem[{Liu et~al.(2019)Liu, Fu, Cao, de~Melo, Tam, Niu, and
  Zhou}]{liu-etal-2019-rhetorically}
Zhiqiang Liu, Zuohui Fu, Jie Cao, Gerard de~Melo, Yik-Cheung Tam, Cheng Niu,
  and Jie Zhou. 2019.
\newblock \href {https://doi.org/10.18653/v1/P19-1192} {Rhetorically controlled
  encoder-decoder for modern {C}hinese poetry generation}.
\newblock In \emph{Proceedings of the 57th Annual Meeting of the Association
  for Computational Linguistics}, pages 1992--2001, Florence, Italy.
  Association for Computational Linguistics.

\bibitem[{Niu et~al.(2017)Niu, Martindale, and Carpuat}]{Niu2017ASO}
Xing Niu, Marianna Martindale, and Marine Carpuat. 2017.
\newblock \href {https://doi.org/10.18653/v1/D17-1299} {A study of style in
  machine translation: Controlling the formality of machine translation
  output}.
\newblock In \emph{Proceedings of the 2017 Conference on Empirical Methods in
  Natural Language Processing}, pages 2814--2819, Copenhagen, Denmark.
  Association for Computational Linguistics.

\bibitem[{Papineni et~al.(2002)Papineni, Roukos, Ward, and
  Zhu}]{papineni2002bleu}
Kishore Papineni, Salim Roukos, Todd Ward, and Wei-Jing Zhu. 2002.
\newblock \href {https://doi.org/10.3115/1073083.1073135} {Bleu: A method for
  automatic evaluation of machine translation}.
\newblock In \emph{Proceedings of the 40th Annual Meeting on Association for
  Computational Linguistics}, ACL '02, pages 311--318, Stroudsburg, PA, USA.
  Association for Computational Linguistics.

\bibitem[{See et~al.(2017)See, Liu, and Manning}]{see2017get}
Abigail See, Peter~J. Liu, and Christopher~D. Manning. 2017.
\newblock \href {https://doi.org/10.18653/v1/P17-1099} {Get to the point:
  Summarization with pointer-generator networks}.
\newblock In \emph{Proceedings of the 55th Annual Meeting of the Association
  for Computational Linguistics (Volume 1: Long Papers)}, pages 1073--1083,
  Vancouver, Canada. Association for Computational Linguistics.

\bibitem[{Sennrich et~al.(2016)Sennrich, Haddow, and Birch}]{Sennrich_2016}
Rico Sennrich, Barry Haddow, and Alexandra Birch. 2016.
\newblock \href {https://doi.org/10.18653/v1/P16-1162} {Neural machine
  translation of rare words with subword units}.
\newblock In \emph{Proceedings of the 54th Annual Meeting of the Association
  for Computational Linguistics (Volume 1: Long Papers)}, pages 1715--1725,
  Berlin, Germany. Association for Computational Linguistics.

\bibitem[{Vaswani et~al.(2017)Vaswani, Shazeer, Parmar, Uszkoreit, Jones,
  Gomez, Kaiser, and Polosukhin}]{vaswani2017attention}
Ashish Vaswani, Noam Shazeer, Niki Parmar, Jakob Uszkoreit, Llion Jones,
  Aidan~N Gomez, {\L}ukasz Kaiser, and Illia Polosukhin. 2017.
\newblock Attention is all you need.
\newblock In \emph{Advances in neural information processing systems}, pages
  5998--6008.

\bibitem[{Wang et~al.(2019{\natexlab{a}})Wang, Hua, and
  Wan}]{Wang-etal-2019-controllable}
Ke~Wang, Hang Hua, and Xiaojun Wan. 2019{\natexlab{a}}.
\newblock \href
  {http://papers.nips.cc/paper/9284-controllable-unsupervised-text-attribute-transfer-via-editing-entangled-latent-representation.pdf}
  {Controllable unsupervised text attribute transfer via editing entangled
  latent representation}.
\newblock In \emph{Advances in Neural Information Processing Systems}, pages
  11034--11044. Curran Associates, Inc.

\bibitem[{Wang et~al.(2019{\natexlab{b}})Wang, Gupta, Chang, and
  Baldridge}]{wang2019task}
Su~Wang, Rahul Gupta, Nancy Chang, and Jason Baldridge. 2019{\natexlab{b}}.
\newblock \href {https://doi.org/10.1609/aaai.v33i01.33017176} {A task in a
  suit and a tie: Paraphrase generation with semantic augmentation}.
\newblock \emph{Proceedings of the AAAI Conference on Artificial Intelligence},
  33:7176–7183.

\bibitem[{Wieting and Gimpel(2018)}]{Wieting_2018}
John Wieting and Kevin Gimpel. 2018.
\newblock \href {https://doi.org/10.18653/v1/P18-1042} {{P}ara{NMT}-50{M}:
  Pushing the limits of paraphrastic sentence embeddings with millions of
  machine translations}.
\newblock In \emph{Proceedings of the 56th Annual Meeting of the Association
  for Computational Linguistics (Volume 1: Long Papers)}, pages 451--462,
  Melbourne, Australia. Association for Computational Linguistics.

\bibitem[{Wiseman et~al.(2018)Wiseman, Shieber, and Rush}]{wiseman2018learning}
Sam Wiseman, Stuart Shieber, and Alexander Rush. 2018.
\newblock \href {https://doi.org/10.18653/v1/D18-1356} {Learning neural
  templates for text generation}.
\newblock In \emph{Proceedings of the 2018 Conference on Empirical Methods in
  Natural Language Processing}, pages 3174--3187, Brussels, Belgium.
  Association for Computational Linguistics.

\bibitem[{Yang et~al.(2019)Yang, Wu, Yang, Xu, and Li}]{yang-etal-2019-low}
Ze~Yang, Wei Wu, Jian Yang, Can Xu, and Zhoujun Li. 2019.
\newblock \href {https://doi.org/10.18653/v1/D19-1197} {Low-resource response
  generation with template prior}.
\newblock In \emph{Proceedings of the 2019 Conference on Empirical Methods in
  Natural Language Processing and the 9th International Joint Conference on
  Natural Language Processing (EMNLP-IJCNLP)}, pages 1886--1897, Hong Kong,
  China. Association for Computational Linguistics.

\bibitem[{Yin et~al.(2019)Yin, Huang, Dai, and Chen}]{Yin-etal-2019-utilizing}
Di~Yin, Shujian Huang, Xin-Yu Dai, and Jiajun Chen. 2019.
\newblock \href {https://doi.org/10.24963/ijcai.2019/747} {Utilizing
  non-parallel text for style transfer by making partial comparisons}.
\newblock In \emph{Proceedings of the Twenty-Eighth International Joint
  Conference on Artificial Intelligence, {IJCAI-19}}, pages 5379--5386.
  International Joint Conferences on Artificial Intelligence Organization.

\bibitem[{Zhang et~al.(2019)Zhang, Yang, Yuan, Shen, and
  Carin}]{zhang2019syntax}
Xinyuan Zhang, Yi~Yang, Siyang Yuan, Dinghan Shen, and Lawrence Carin. 2019.
\newblock \href {https://doi.org/10.18653/v1/P19-1199} {Syntax-infused
  variational autoencoder for text generation}.
\newblock In \emph{Proceedings of the 57th Annual Meeting of the Association
  for Computational Linguistics}, pages 2069--2078, Florence, Italy.
  Association for Computational Linguistics.

\end{thebibliography}
